\pdfoutput=1

\documentclass[11pt]{article}
\usepackage[]{authblk}
\usepackage{blindtext}

\usepackage[]{ACL2023}

\usepackage{times}
\usepackage{latexsym}
\usepackage{enumitem}

\usepackage[T1]{fontenc}

\usepackage[utf8]{inputenc}

\usepackage{microtype}
\usepackage{graphicx}
\usepackage{multirow}
\usepackage{amsmath}
\usepackage{dashbox}

\usepackage{fbox}
\usepackage{longfbox}
\usepackage{amssymb}
\usepackage{gb4e}
\usepackage{booktabs}
\usepackage{tabularx}
\usepackage{color}
\usepackage{array}
\usepackage{caption}
\usepackage{subcaption}
\usepackage{fontawesome}
\usepackage{graphicx}
\usepackage{blindtext}

\newcolumntype{Z}{>{\raggedright\let\newline\\\arraybackslash\hspace{0pt}}X}
\noautomath

\newcommand*{\email}[1]{%
    \small\href{mailto:#1}{#1}\par
    }

\title{Incorporating Singletons and Mention-based Features in Coreference Resolution via Multi-task Learning for Better Generalization}

\usepackage{fdsymbol}
\author[$\spadesuit$]{\textbf{Yilun Zhu}}
\author[$\heartsuit$]{\textbf{Siyao Peng}}
\author[$\clubsuit$$\diamondsuit$]{\textbf{Sameer Pradhan}}
\author[$\spadesuit$]{\textbf{Amir Zeldes}}
\affil[$\spadesuit$]{Department of Linguistics, Georgetown University}
\affil[$\heartsuit$]{Center for Information and Language Processing (CIS), LMU Munich}
\affil[$\clubsuit$]{Linguistic Data Consortium, University of Pennsylvania}
\affil[$\diamondsuit$]{cemantix.org}
\affil[ ]{\tt \email{\{yz565, Amir.Zeldes\}@georgetown.edu, siyaopeng@cis.lmu.de, pradhan@cemantix.org}}

\begin{document}
\maketitle
\begin{abstract}
Previous attempts to incorporate a mention detection step into end-to-end neural coreference resolution for English have been hampered by the lack of singleton mention span data as well as other entity information. This paper presents a coreference model that learns singletons as well as features such as entity type and information status via a multi-task learning-based approach. This approach achieves new state-of-the-art scores on the OntoGUM benchmark (+2.7 points) and increases robustness on multiple out-of-domain datasets (+2.3 points on average), likely due to greater generalizability for mention detection and utilization of more data from singletons when compared to only coreferent mention pair matching.\footnote{The code is publicly available at \url{https://github.com/yilunzhu/coref-mtl}.}
\end{abstract}


\section{Introduction} \label{sec:intro}
Coreference is a linguistic phenomenon that occurs when two or more expressions in a text refer to the same entity (e.g.~\textit{the Vice President... She}). Conceptually, resolving coreference takes two steps: identifying all mention candidates from a text as opposed to non-referring expressions, and linking identified mentions into clusters. However, in a given document, some mentions are never referred back to: these are called singletons, i.e.~mentions that, unlike non-referring expressions, 
could be referred back to in principle, but are not involved in any coreference relations in context. Singletons are important to coreference resolution since they represent true negatives in cluster linking \citep{kubler-zhekova-2011-singletons}, but also to how humans understand discourse from a theoretical perspective \citep{grosz-etal-1995-centering}, since they also constitute mentioned entities (i.e.~clusters of size 1).

However, due to the lack of singleton annotation in the most frequently used coreference dataset for English, i.e.~OntoNotes V5.0 \citep{weischedel-handbook-2011-notes,pradhan-etal-2013-towards}, previous attempts have either ignored singletons \citep{lee-etal-2017-end, lee-etal-2018-higher, wu-etal-2020-corefqa, dobrovolskii-2021-word} or incorporated pseudo-singletons into the model \citep{wu-gardner-2021-understanding, toshniwal-etal-2021-generalization}.
The first approach is commonly used in contemporary end-to-end (e2e) systems which train directly on detecting coreferring mentions, but causes problems in that models cannot differentiate singleton spans from non-referring or random/meaningless spans, i.e.~penalizing these two types equally. 
Though e2e has achieved significant progress on OntoNotes, it does not align with linguistic theories on how humans resolve the task. The second approach attempts to amend the model with pseudo-singletons by predicting non-coreferring mentions, but the accuracy gap between gold and generated singletons is unknown and ultimately leads to degradation.


Previous work has also shown that recent coreference models struggle with domain generalization \citep{moosavi-strube-2017-lexical, zhu-etal-2021-ontogum}. To alleviate the problem, \citet{moosavi-strube-2018-using} proposed a novel algorithm to incorporate linguistic features and showed improvement in out-of-domain (OOD) data. \citet{subramanian-roth-2019-improving} applied adversarial
training to improve generalization. However, the first approach requires carefully designed linguistic features, and both papers evaluated generalization only on one single-genre dataset, limiting the validity of the results.

To tackle these challenges, we introduce a novel coreference model. Our contributions can be summarized as follows: First, we propose 
a multi-task learning (MTL) based neural coreference model with constrained mention detection, which jointly learns several mention-based tasks, including singleton detection, entity type recognition, and information status classification. Second, experiments demonstrate that the proposed model achieves new state-of-the-art performance on the OntoGUM test set. Third, we show that our model outperforms strong baselines on two OOD datasets, showing it generalizes more reliably to unseen data than plain e2e. We release all code and provide a system that detects and links all mentions, including singletons, and outputs predicted entity types. 



\section{Related Work} \label{sec:literature}
\paragraph{MTL for coreference}
Multitask learning \cite{Caruana1997,CollobertWeston2008} uses a single model with shared parameters trained to perform multiple tasks, with potential benefits arising from synergies between related objectives.
Previous work has investigated the use of MTL for coreference by harnessing related pre-training tasks. 
\citet{yu-poesio-2020-multitask, kobayashi-etal-2022-constrained} applied an MTL framework to a more specific bridging resolution problem, with standard coreference resolution as the additional task. \citet{luan-etal-2018-multi} used MTL with coreference resolution, entity recognition, and relation extraction for scientific knowledge graph construction. \citet{lu-ng-2021-constrained} used five MTL tasks for event coreference resolution. 

\paragraph{Neural coreference resolution}
The e2e approach jointly learns mention detection and coreferent pair scoring \citep{lee-etal-2017-end}, and achieved SOTA scores on the OntoNotes test set before several extensions were proposed. \citet{lee-etal-2018-higher, kantor-globerson-2019-coreference} improved span representations to improve pair matching. 
\citet{joshi-etal-2020-spanbert} added better pre-trained language models to gain additional score boosting. \citet{wu-etal-2020-corefqa} adapted a question-answering framework into the task and improved both span detection and coreference matching scores. \citet{dobrovolskii-2021-word} also improved performance by initially matching coreference links via words instead of spans.

\section{Methods} \label{sec:model}
\subsection{Model}
Let $N$ be the number of possible spans in a document $D$. The coreference task can be formulated as assigning an antecedent span $y_i$ for each span $i$, where the set of possible antecedents for each span $i$ contains a dummy antecedent $\epsilon$ and all preceding spans: $\mathcal{Y}(i) = \{\epsilon, 1,...,i-1\}$. 

\vspace{-3mm}
\[s(i, j) = 
\begin{cases}
0, & j = \epsilon\\
s_m(i) + s_m(j) + s_c(i, j), & j \neq \epsilon
\end{cases}\]

\noindent where $s_m(i)$ and $s_m(j)$ are the mention scores that determine how likely the selected text span is a mention candidate. 
Previous work utilizes a scoring function to measure how likely the span is a coreference markable. However, singletons in the training data are ignored and thus weaken the model's generalization capability. Therefore, our proposed model uses two scoring functions to represent the distributions of markables and mentions better. The mention scoring function uses two feed-forward networks fed by the representation of each span: one part is a markable score that calculates the score of the span being a coreferent markable in the document; the other is the mention candidate score that determines how likely a span is a mention candidate. The formula is represented as follows:

\vspace{-3mm}
\[s_m(i) = \beta_1 \cdot s_{markable(i)} + \beta_2 \cdot s_{mention(i)}\]
\[s_{markable(i)} = w_{markble} \cdot \textsc{ffnn}(g_i) \]
\[s_{mention(i)} = w_{mention} \cdot \textsc{ffnn}(g_i) \]

\noindent where $\cdot$ denotes a dot product, $\textsc{ffnn}$ denotes a feed-forward neural network, $\beta_1$ and $\beta_2$ denote model parameters that adjust the weights of markable scores and mention candidate scores, and $g_i$ denotes the represented embeddings of the span (we use the same span representing method as in \citet{lee-etal-2017-end}). The two scoring functions are computed via two standard feed-forward neural networks. The purpose of this design is to prevent random text spans being fed to the pair-matching step. Following the e2e approach \citep{lee-etal-2017-end, lee-etal-2018-higher, joshi-etal-2020-spanbert}, we concatenate the boundary representations, the soft head vector and an additional feature vector $\phi$ containing speaker information, and feed the resulting vector into separate feed-forward neural networks to calculate markable scores and mention candidate scores.

In addition to the main pair-matching task, our model adds three mention-based tasks: a (possibly singleton) mention span detection task, entity type recognition, and information status classification (see below). For each task, the span vector is fed into a separate feed-forward network for classification. Each task is assigned a weight to calculate the total loss score:

\vspace{-3mm}
\[\mathcal{L}_{total} = \sum_{c=1}^C \mathcal{W}_c \cdot \mathcal{L}_c\]

\noindent where $\mathcal{W}_c$ is the weight for task $c$. See Appendix \ref{sec:appendix_modelarc} for an overview of the model architecture.

\begin{table*}[t!hb]
    \centering\small
    \resizebox{\textwidth}{!}{%
    \begin{tabular}{l|ccc|ccccccccc|c}
    \toprule
    \multicolumn{1}{l|}{\multirow{2}{*}{}} & \multicolumn{3}{c|}{Markble Detection} & \multicolumn{3}{c}{MUC} & \multicolumn{3}{c}{B$^3$} & \multicolumn{3}{c|}{CEAF$_{\phi4}$} & \multicolumn{1}{c}{\multirow{2}{*}{Avg. F1}} \\
    \multicolumn{1}{l|}{} & P & R & \multicolumn{1}{c|}{F1} & P & R & F1 & P & R & F1 & P & R & F1 &  \\
    \hline
    \multicolumn{14}{l}{\rule{0pt}{2ex}\textbf{In-domain} - \textsc{OntoGUM}}\\
    \hline
    \rule{0pt}{2ex}\citet{DBLP:journals/corr/abs-1907-10529} & \textbf{91.0} & 71.9 & 80.3 & \textbf{83.3} & 69.7 & 75.9 & 70.8 & 59.2 & 64.5 & 70.5 & 45.8 & 55.5 & 65.5 \\
    \rule{0pt}{2ex}MTL (sg) & 90.2 & 75.0 & \textbf{81.9} & 82.7 & 72.8 & 77.4 & 70.4 & 63.1 & 66.5 & 71.5 & 49.2 & 58.3 & 67.6 \\
    \rule{0pt}{2ex}MTL (sg\texttt{+}ent) & 90.0 & \textbf{75.1} & \textbf{81.9} & 82.8 & \textbf{72.9} & \textbf{77.6} & \textbf{71.2} & \textbf{63.6} & \textbf{67.2} & \textbf{71.9} & \textbf{50.2} & \textbf{59.1} & \textbf{68.2} \\
    \rule{0pt}{2ex}MTL (sg\texttt{+}ent\texttt{+}infs.) & 90.0 & 75.0 & 81.8 & 82.1 & 72.3 & 76.9 & 70.0 & 62.3 & 65.9 & 70.0 & 48.6 & 57.3 & 66.9 \\
    
    \midrule\midrule
    \multicolumn{14}{l}{\textbf{Out-of-domain} - \textsc{OntoNotes}}\\
    \hline
    \rule{0pt}{2ex}\citet{DBLP:journals/corr/abs-1907-10529} & \textbf{83.9} & 76.9 & 80.3 & \textbf{77.6} & 72.7 & 75.1 & 66.9 & 60.6 & 63.6 & \textbf{64.3} & 54.5 & 59.0 & 65.9 \\
    \rule{0pt}{2ex}MTL (sg\texttt{+}ent) & 82.2 & \textbf{80.2} & \textbf{81.2} & 77.0 & \textbf{76.1} & \textbf{76.5} & \textbf{67.1} & \textbf{64.0} & \textbf{65.5} & 63.6 & \textbf{59.5} & \textbf{61.5} & \textbf{67.8} \\
    
    \hline
    \multicolumn{14}{l}{\rule{0pt}{2ex}\textbf{Out-of-domain} - \textsc{WikiCoref}}\\ 
    \hline
    \rule{0pt}{2ex}\citet{DBLP:journals/corr/abs-1907-10529} & 79.9 & 58.8 & 67.7 & 73.7 & 60.1 & 66.2 & 66.4 & 43.4 & 52.4 & 56.6 & 31.6 & 40.5 & 53.0 \\
    \rule{0pt}{2ex}MTL (sg\texttt{+}ent) & \textbf{80.4} & \textbf{60.0} & \textbf{68.7} & \textbf{74.5} & \textbf{61.8} & \textbf{67.5} & \textbf{67.8} & \textbf{45.3} & \textbf{54.4} & \textbf{59.0} & \textbf{33.0} & \textbf{42.4} & \textbf{55.6} \\
    
    \bottomrule
    \end{tabular}
    }
    \caption{Comparison between \citet{DBLP:journals/corr/abs-1907-10529} and our model on test sets of both in-domain (OntoGUM 8.0) and out-of-domain datasets (OntoNotes and WikiCoref). The overall F1 score is the average of F1s from three evaluation metrics MUC, B$^3$, and CEAF$_{\phi4}$. All models are trained on OntoGUM.}
    \label{tab:res}
\end{table*}

\subsection{Task Selection}
Since OntoNotes does not contain singletons, we choose a corpus for which singleton information is available but follows the same annotation scheme as OntoNotes. The OntoGUM corpus \citep{zhu-etal-2021-ontogum} 
is an adapted version of the GUM corpus \cite{Zeldes2017}, a multi-layer corpus with a range of annotations at the word level (part-of-speech, morphology), phrase level (phrase trees, entity recognition, and linking), dependency level (Universal Dependencies syntax) and document-level (discourse parses and coreference). Although OntoGUM uses the same singleton-free coreference scheme as OntoNotes, information about singletons can be recovered from the original GUM corpus. We therefore select three annotations from GUM and investigate whether they are helpful for coreference resolution on OntoGUM: nested mention span detection, entity type, and information status.

\vspace{-0.2em}
\paragraph{Mention detection} As outlined in Section \ref{sec:intro}, we integrate gold nested mentions, including singletons (sg), into our model to improve mention detection and coreference. The task aims to recognize meaningful referential text spans and makes more information available to the model than the plain e2e approach that only trains on coreferring mentions ($\sim$39\% of mentions in GUM are singletons). 

\vspace{-0.2em}
\paragraph{Entity type} GUM assigns one of ten entity types (ent) to each mention -- person, organization, etc. (see Figure \ref{fig:og-all} in Appendix \ref{sec:appendix_sample}). Since a cluster usually has one entity type, this feature instructs the model regarding which mentions belong to the same semantic class.

\vspace{-0.2em}
\paragraph{Information status} Information status (infs.) indicates how an entity was introduced into discourse, e.g.~new, previously mentioned or inferrable from other mentions \citep{Prince_Taxonomy81}. Each mention is assigned one of six labels (see Appendix \ref{sec:appendix_infstat}).
This task is expected to inform the model about the likelihood and how an entity was previously introduced.

\section{Experiments} \label{sec:experiments}

\subsection{Datasets}
OntoGUM \citep{zhu-etal-2021-ontogum} is a coreference dataset following the same annotation scheme as OntoNotes.  This paper adds other layers to the coreference annotation, such as mention spans (including singletons), aligned entity types, and information status,  automatically extracted from the GUM corpus. We train the model with GUM v8.0, which includes 193 documents across 12 written and spoken genres with $\sim$180K tokens. 

We also evaluate our model on two OOD datasets of the same annotation scheme: OntoNotes and WikiCoref. OntoNotes includes richly annotated documents with layers including syntax, propositions, named entities, word senses, and coreference, but no singleton mentions or aligned (non-named) entity types \citep{pradhan-etal-2013-towards}. Its test set includes 348 documents with 170K tokens. WikiCoref \citep{ghaddar-langlais-2016-wikicoref} is a manually annotated corpus from English Wikipedia, containing 30 documents with $\sim$60K tokens. 


\subsection{Baseline}
Combining the e2e approach with a contextualized language model (LM) and span masking is one of the best models on OntoNotes. Following \citet{joshi-etal-2020-spanbert}, we use large SpanBERT embeddings as the LM and the improved coarse-to-fine \citep{lee-etal-2018-higher} SOTA model as our baseline model (see Appendix \ref{sec:appendix_implementation} for implementation details).


\subsection{Task Weights}
The task weights are a list of parameters that controls the relative importance of various tasks in our model, which are optimized via hyperparameter search on the OntoGUM dev set to achieve the best performance. In the optimal setting with 2 auxiliary tasks, the loss weight for the major task coreference relation identification is set to 0.4 and the weights for singleton detection and entity type recognition 
are set to 0.2 each. The weights are 0.15 for each auxiliary task when information status is added to training.

\subsection{Results}
\paragraph{In-domain Evaluation}
We train the model on OntoGUM and evaluate it in-domain.
As shown in the first part of Table \ref{tab:res}, our model with the best setting improves average F1 by 2.7 points and achieves new SOTA performance on the OntoGUM benchmark, indicating the benefit of the MTL tasks. We also note that recall scores of both mention detection and coreference matching show a significant increase by 3.2 and 4.0 points, respectively,
which suggests that the MTL approach helps the model capture more non-trivial markable spans and coreference relations than the baseline model, with little or no precision cost. In addition, though information status contributes to the result as a sole auxiliary task (see Table \ref{tab:abalation}), it is harmful when training with other tasks.

\paragraph{Out-of-domain Evaluation}
To test the robustness of our model, we evaluate on two OOD datasets sharing the same annotation scheme with OntoGUM. The second part of Table \ref{tab:res} shows that our best in-domain model with mention detection and entity type as auxiliary tasks outperforms the baseline model on both datasets by 2.3 points on average. For OntoNotes, though our model has slightly lower precision, the recall results in substantially better performance; for WikiCoref, our model performs better on both precision and recall. These results indicate that the knowledge gained from the multiple mention-based tasks can be transferred to unseen text types, 
and is likely a combination of more training data (since singletons include instances not considered by the baseline training) and the learning of features distinguishing non-mentions from mentions and ones corresponding to semantic types.

\subsection{Ablation Study}
To show the importance of each task in our model, we ablate each task in the architecture and report the average F1 on the OntoGUM development set. In Table \ref{tab:abalation}, singleton scores and the mention detection task contribute 1.3 points to the final result, indicating that this feature is the most important one.

\begin{table}[t!hb]
    \centering\small
    \begin{tabular}{p{4.5cm}cc}
    \toprule
     & Avg. F1 & $\Delta$ \\
    \hline
    \rule{0pt}{2ex}Base model & 67.0 & \\
    \hspace{1ex}w/ singleton detection (=sg) & 68.3 & +1.3\\
    \hspace{1ex}w/ sg + entity type (=et) & 68.7 & +0.4 \\
    \hspace{1ex}w/ sg + et + information status & 67.8 & -0.9 \\
    \bottomrule
    \end{tabular}
    \caption{Comparison of various tasks included in the coreference model on the OntoGUM development data.}
    \label{tab:abalation}
\end{table}


With the addition of the nested entity type recognition task, the model brings a smaller increase (0.4 points) to the final result. There could be several reasons for this: one is that the LM has already learned entity types latently, so giving this as an explicit feature is redundant; the other reason is that the baseline model rarely groups mentions with different entity types into clusters so that entity type features can only correct few errors.

When only integrating information status into the model, the result (avg. F1 67.6) outperforms the baseline model, showing the effectiveness of this type of information. However, when all three tasks are incorporated, the overall score (67.8) is lower than excluding information status classification (68.7), which shows that information status is redundant when other mention-based features are specified.

\section{Error analysis}
\label{sec:appendix_error}
We conduct quantitative and qualitative error analyses to illustrate how our model differs from the baseline. 
Firstly we conduct a quantitative analysis following \citet{lu-ng-2020-conundrums}, who classify resolution errors into 13 classes. Following their approach, we merge coreference errors into 6 groups. Table \ref{tab:quant_error} displays the distribution of errors observed in the OntoGUM development set. These errors are present in the baseline e2e model but correctly resolved by our proposed MTL model (e2e errors) or vice versa (mtl errors).

\begin{table}[t!bh]
\centering\small
\resizebox{0.48\textwidth}{!}{%
    \begin{tabular}{l|cr|cr}
        \toprule
        Error type & \multicolumn{2}{c|}{mtl errors} & \multicolumn{2}{c}{e2e errors} \\
        \hline
        \rule{0pt}{2ex}Pronouns & & \\
        \hspace{4pt}- 1st \& 2nd person pronouns & 6 & 3.6\% & 12 & 5.0\% \\
        \hspace{4pt}- 3rd person pronouns & 20 &  12.1\% & 68 & 28.3\% \\
        Definiteness & & \\
        \hspace{4pt}- Definite nouns & 63 & 38.2\% & 98 & 40.8\% \\
        \hspace{4pt}- Indefinite nouns & 13 & 7.9\% & 13 & 5.4\% \\
        Proper nouns & 23 & 14.0\% & 19 & 7.9\% \\
        Others & 40 & 24.2\% & 30 & 12.5\% \\
        \hline
        \rule{0pt}{2ex}Total & 165 & 100.0\% & 240 & 100.0\% \\
        \bottomrule
    \end{tabular}
}
    \caption{Number and percentage of errors by class that are produced by e2e but avoided by the MTL model (e2e errors) and produced by the MTL model but resolved by the e2e model (mtl errors).}
    \label{tab:quant_error}
\end{table}

The majority of mtl errors involve definite nominals, revealing the challenge of resolving cherry-picked cases that must be memorized within a multi-genre context. However, our proposed model demonstrates its ability to correctly identify relations when multiple clusters are involved. Furthermore, nearly 16\% of resolved errors are associated with pronouns, indicating that our model is more capable of accurately identifying coreference relationships within the context of third-person pronouns and demonstrates a slight improvement in handling pronouns in dialogue, particularly first and second-person pronouns.

We also observe that our proposed model reduces errors across nearly all types compared to the baseline model, particularly in the case of third-person pronouns. This result suggests that integrating entity type recognition and mention detection in the MTL framework enables accurate recognition of noun-pronoun relations, particularly for pronouns that do not provide explicit entity type information, e.g., \textit{it}. Additionally, the MTL model demonstrates improved error avoidance with definite nouns. These findings highlight the enhanced performance of our proposed model in identifying coreference relations within the local context.

\begin{table}[t!bh]
    \centering\small
    \begin{tabularx}{0.48\textwidth}{>{\hsize=.01\hsize}XZ}
    \toprule
     \multicolumn{2}{l}{\textbf{Entity type errors}} \\
      \rule{0pt}{2ex}1 & he did represent [\textcolor{red}{the school}]$_1$ during the very first Eton v [\textcolor{blue}{Harrow}]$_1$ cricket match \\ 
      2 & Who cut [\textcolor{red}{the grass}]$_1$? Marlena did [\textcolor{red}{it}]$_2$. Marlena did [\textcolor{red}{it}]$_2$ a long time ago, but [it]$_1$ hasn’t been watered. [It]$_1$’s dying. \\ 
      3 & I made [\textcolor{blue}{noises}]$_1$ with \textcolor{red}{my heels} but [\textcolor{red}{they}]$_1$ were too loud so I stopped. \\ 
      \midrule
      \multicolumn{2}{l}{\rule{0pt}{2ex}\textbf{Singleton errors}} \\ 
      \rule{0pt}{2.2ex}4 & The main reason attributed for the pollution of Athens is because the city is enclosed by mountains in [\textcolor{red}{a basin which does not let the smog leave}]$_1$ ... have greatly contributed to better atmospheric conditions in [\textcolor{blue}{the basin}]$_1$. \\ 
      5 & This means that if [\textcolor{red}{the govt}]$_1$ decided to print 1 quadrillion dollars in the span of a week ... we 're loaning [\textcolor{red}{the US govt}]$_1$ \textcolor{red}{the very money it prints} \\ 
    \bottomrule
    \end{tabularx}
    \caption{A qualitative analysis of OntoGUM dev errors that appear in the e2e model but are avoided by our MTL model. MTL predictions (gold) are represented by [brackets]$_x$. E2e predictions (errors) are highlighted in colored text and each color in an example denotes a coreference cluster.}
    \label{tab:error}
    \vspace{-1em}
\end{table}

We also identify several errors that illustrate the impact of singleton detection and entity type recognition. Examples in Table \ref{tab:error} demonstrate how including singletons and mention-based features improves the retrieval of accurate mention spans and enhances coreference relationships. The first three examples highlight how entity-type recognition contributes to resolution by avoiding type mismatches. 
In example (1), the pressure from entity type recognition likely aids in identifying \textit{Harrow} as a school (an \textsc{organization}). 
In example (2), the MTL model recognizes \textit{it} as an \textsc{event}, thereby correctly creating two distinct groups and avoiding coreference with \textit{the grass} (a \textsc{plant} entity). Similarly, example (3) presents pressure to recognize that \textit{they} is not an inanimate \textsc{object}, so it correctly prefers \textit{noises} as the antecedent. Examples (4) and (5) illustrate how mention detection identifies missing mentions in the baseline model or improves boundary recognition. These representative examples provide valuable insights into the significance of incorporating singletons and auxiliary mention-based tasks into a coreference model.

\section{Conclusion} \label{sec:conclusion}
This paper presents a neural coreference model that connects singletons and other mention-based features to coreference relation matching via an MTL architecture, which (1) outperforms a strong baseline and achieves new SOTA results on OntoGUM and (2) beats the baseline model on two unseen datasets. The results show the effect of singletons and mention features and indicate improvements in model robustness when transferring to unseen data rather than overfitting distributions in the training data. In addition, our resulting system can output all mentions (incl. singletons) with entity types out-of-the-box, which benefits a series of downstream applications such as Entity Linking, Dialogue Systems, Machine Translation, Summarization, and more, since our single model already outputs typed spans for all entities mentioned in a text (see Figure \ref{fig:og-all-mtl} in Appendix \ref{sec:appendix_sample} for an illustration).

\section*{Limitations}
In this work, we have experimented with training our model on OntoGUM. Due to the lack of singletons and other mention-based annotations, we do not train the model on the most frequently used and one of the largest coreference datasets. Thus the proposed model has not been tested on a large-scale dataset and compared with other coreference models on OntoNotes.

We evaluate the model on two English OOD datasets to investigate the model generalization. Several coreference datasets in other languages share the same annotation scheme as OntoGUM, such as Arabic \citep{pradhan-etal-2013-towards}, and Chinese \citep{pradhan-etal-2013-towards}. The proposed model needs to be evaluated on datasets in other languages and demonstrate the model generalization across languages. However, this would require singleton annotated data in those languages as well. With recent releases such as CorefUD \cite{nedoluzhko-etal-2022-corefud} promoting standardization of multilingual coreference annotations and singleton annotations, we are hopeful that such experiments will be possible in the near future.

\bibliography{anthology,custom}

\begin{thebibliography}{30}
\expandafter\ifx\csname natexlab\endcsname\relax\def\natexlab#1{#1}\fi

\bibitem[{Caruana(1997)}]{Caruana1997}
Rich Caruana. 1997.
\newblock \href {https://doi.org/10.1023/A:1007379606734} {Multitask learning}.
\newblock \emph{Mach. Learn.}, 28(1):41–75.

\bibitem[{Collobert and Weston(2008)}]{CollobertWeston2008}
Ronan Collobert and Jason Weston. 2008.
\newblock \href {https://doi.org/10.1145/1390156.1390177} {A unified architecture for natural language processing: Deep neural networks with multitask learning}.
\newblock In \emph{Proceedings of the 25th International Conference on Machine Learning}, ICML '08, pages 160–--167, New York, NY, USA. Association for Computing Machinery.

\bibitem[{Dobrovolskii(2021)}]{dobrovolskii-2021-word}
Vladimir Dobrovolskii. 2021.
\newblock \href {https://doi.org/10.18653/v1/2021.emnlp-main.605} {Word-level coreference resolution}.
\newblock In \emph{Proceedings of the 2021 Conference on Empirical Methods in Natural Language Processing}, pages 7670--7675, Online and Punta Cana, Dominican Republic. Association for Computational Linguistics.

\bibitem[{Ghaddar and Langlais(2016)}]{ghaddar-langlais-2016-wikicoref}
Abbas Ghaddar and Phillippe Langlais. 2016.
\newblock \href {https://aclanthology.org/L16-1021} {{W}iki{C}oref: An {E}nglish coreference-annotated corpus of {W}ikipedia articles}.
\newblock In \emph{Proceedings of the Tenth International Conference on Language Resources and Evaluation ({LREC}'16)}, pages 136--142, Portoro{\v{z}}, Slovenia. European Language Resources Association (ELRA).

\bibitem[{Grosz et~al.(1995)Grosz, Joshi, and Weinstein}]{grosz-etal-1995-centering}
Barbara~J. Grosz, Aravind~K. Joshi, and Scott Weinstein. 1995.
\newblock \href {https://aclanthology.org/J95-2003} {{C}entering: A framework for modeling the local coherence of discourse}.
\newblock \emph{Computational Linguistics}, 21(2):203--225.

\bibitem[{Hou(2020)}]{hou-2020-bridging}
Yufang Hou. 2020.
\newblock \href {https://doi.org/10.18653/v1/2020.acl-main.132} {Bridging anaphora resolution as question answering}.
\newblock In \emph{Proceedings of the 58th Annual Meeting of the Association for Computational Linguistics}, pages 1428--1438, Online. Association for Computational Linguistics.

\bibitem[{Joshi et~al.(2019)Joshi, Chen, Liu, Weld, Zettlemoyer, and Levy}]{DBLP:journals/corr/abs-1907-10529}
Mandar Joshi, Danqi Chen, Yinhan Liu, Daniel~S. Weld, Luke Zettlemoyer, and Omer Levy. 2019.
\newblock \href {http://arxiv.org/abs/1907.10529} {{SpanBERT}: Improving pre-training by representing and predicting spans}.
\newblock \emph{CoRR}, abs/1907.10529.

\bibitem[{Joshi et~al.(2020)Joshi, Chen, Liu, Weld, Zettlemoyer, and Levy}]{joshi-etal-2020-spanbert}
Mandar Joshi, Danqi Chen, Yinhan Liu, Daniel~S. Weld, Luke Zettlemoyer, and Omer Levy. 2020.
\newblock \href {https://doi.org/10.1162/tacl_a_00300} {{S}pan{BERT}: Improving pre-training by representing and predicting spans}.
\newblock \emph{Transactions of the Association for Computational Linguistics}, 8:64--77.

\bibitem[{Kantor and Globerson(2019)}]{kantor-globerson-2019-coreference}
Ben Kantor and Amir Globerson. 2019.
\newblock \href {https://doi.org/10.18653/v1/P19-1066} {Coreference resolution with entity equalization}.
\newblock In \emph{Proceedings of the 57th Annual Meeting of the Association for Computational Linguistics}, pages 673--677, Florence, Italy. Association for Computational Linguistics.

\bibitem[{Kobayashi et~al.(2022)Kobayashi, Hou, and Ng}]{kobayashi-etal-2022-constrained}
Hideo Kobayashi, Yufang Hou, and Vincent Ng. 2022.
\newblock \href {https://doi.org/10.18653/v1/2022.acl-long.56} {Constrained multi-task learning for bridging resolution}.
\newblock In \emph{Proceedings of the 60th Annual Meeting of the Association for Computational Linguistics (Volume 1: Long Papers)}, pages 759--770, Dublin, Ireland. Association for Computational Linguistics.

\bibitem[{K{\"u}bler and Zhekova(2011)}]{kubler-zhekova-2011-singletons}
Sandra K{\"u}bler and Desislava Zhekova. 2011.
\newblock \href {https://aclanthology.org/R11-1036} {Singletons and coreference resolution evaluation}.
\newblock In \emph{Proceedings of the International Conference Recent Advances in Natural Language Processing 2011}, pages 261--267, Hissar, Bulgaria. Association for Computational Linguistics.

\bibitem[{Lee et~al.(2017)Lee, He, Lewis, and Zettlemoyer}]{lee-etal-2017-end}
Kenton Lee, Luheng He, Mike Lewis, and Luke Zettlemoyer. 2017.
\newblock \href {https://doi.org/10.18653/v1/D17-1018} {End-to-end neural coreference resolution}.
\newblock In \emph{Proceedings of the 2017 Conference on Empirical Methods in Natural Language Processing}, pages 188--197, Copenhagen, Denmark. Association for Computational Linguistics.

\bibitem[{Lee et~al.(2018)Lee, He, and Zettlemoyer}]{lee-etal-2018-higher}
Kenton Lee, Luheng He, and Luke Zettlemoyer. 2018.
\newblock \href {https://doi.org/10.18653/v1/N18-2108} {Higher-order coreference resolution with coarse-to-fine inference}.
\newblock In \emph{Proceedings of the 2018 Conference of the North {A}merican Chapter of the Association for Computational Linguistics: Human Language Technologies, Volume 2 (Short Papers)}, pages 687--692, New Orleans, Louisiana. Association for Computational Linguistics.

\bibitem[{Lu and Ng(2020)}]{lu-ng-2020-conundrums}
Jing Lu and Vincent Ng. 2020.
\newblock \href {https://doi.org/10.18653/v1/2020.emnlp-main.536} {Conundrums in entity coreference resolution: Making sense of the state of the art}.
\newblock In \emph{Proceedings of the 2020 Conference on Empirical Methods in Natural Language Processing (EMNLP)}, pages 6620--6631, Online. Association for Computational Linguistics.

\bibitem[{Lu and Ng(2021)}]{lu-ng-2021-constrained}
Jing Lu and Vincent Ng. 2021.
\newblock \href {https://doi.org/10.18653/v1/2021.naacl-main.356} {Constrained multi-task learning for event coreference resolution}.
\newblock In \emph{Proceedings of the 2021 Conference of the North American Chapter of the Association for Computational Linguistics: Human Language Technologies}, pages 4504--4514, Online. Association for Computational Linguistics.

\bibitem[{Luan et~al.(2018)Luan, He, Ostendorf, and Hajishirzi}]{luan-etal-2018-multi}
Yi~Luan, Luheng He, Mari Ostendorf, and Hannaneh Hajishirzi. 2018.
\newblock \href {https://doi.org/10.18653/v1/D18-1360} {Multi-task identification of entities, relations, and coreference for scientific knowledge graph construction}.
\newblock In \emph{Proceedings of the 2018 Conference on Empirical Methods in Natural Language Processing}, pages 3219--3232, Brussels, Belgium. Association for Computational Linguistics.

\bibitem[{Moosavi and Strube(2017)}]{moosavi-strube-2017-lexical}
Nafise~Sadat Moosavi and Michael Strube. 2017.
\newblock \href {https://doi.org/10.18653/v1/P17-2003} {Lexical features in coreference resolution: To be used with caution}.
\newblock In \emph{Proceedings of the 55th Annual Meeting of the Association for Computational Linguistics (Volume 2: Short Papers)}, pages 14--19, Vancouver, Canada. Association for Computational Linguistics.

\bibitem[{Moosavi and Strube(2018)}]{moosavi-strube-2018-using}
Nafise~Sadat Moosavi and Michael Strube. 2018.
\newblock \href {https://doi.org/10.18653/v1/D18-1018} {Using linguistic features to improve the generalization capability of neural coreference resolvers}.
\newblock In \emph{Proceedings of the 2018 Conference on Empirical Methods in Natural Language Processing}, pages 193--203, Brussels, Belgium. Association for Computational Linguistics.

\bibitem[{Nedoluzhko et~al.(2022)Nedoluzhko, Nov{\'a}k, Popel, {\v{Z}}abokrtsk{\'y}, Zeldes, and Zeman}]{nedoluzhko-etal-2022-corefud}
Anna Nedoluzhko, Michal Nov{\'a}k, Martin Popel, Zden{\v{e}}k {\v{Z}}abokrtsk{\'y}, Amir Zeldes, and Daniel Zeman. 2022.
\newblock \href {https://aclanthology.org/2022.lrec-1.520} {{C}oref{UD} 1.0: Coreference meets {U}niversal {D}ependencies}.
\newblock In \emph{Proceedings of the Thirteenth Language Resources and Evaluation Conference}, pages 4859--4872, Marseille, France. European Language Resources Association.

\bibitem[{Pradhan et~al.(2013)Pradhan, Moschitti, Xue, Ng, Bj{\"o}rkelund, Uryupina, Zhang, and Zhong}]{pradhan-etal-2013-towards}
Sameer Pradhan, Alessandro Moschitti, Nianwen Xue, Hwee~Tou Ng, Anders Bj{\"o}rkelund, Olga Uryupina, Yuchen Zhang, and Zhi Zhong. 2013.
\newblock \href {https://aclanthology.org/W13-3516} {Towards robust linguistic analysis using {O}nto{N}otes}.
\newblock In \emph{Proceedings of the Seventeenth Conference on Computational Natural Language Learning}, pages 143--152, Sofia, Bulgaria. Association for Computational Linguistics.

\bibitem[{Prince(1981)}]{Prince_Taxonomy81}
Ellen~F. Prince. 1981.
\newblock Toward a taxonomy of given-new information.
\newblock In P.~Cole, editor, \emph{Syntax and semantics: Vol. 14. Radical Pragmatics}, pages 223--255. Academic Press, New York.

\bibitem[{Roesiger et~al.(2018)Roesiger, Riester, and Kuhn}]{roesiger-etal-2018-bridging}
Ina Roesiger, Arndt Riester, and Jonas Kuhn. 2018.
\newblock \href {https://aclanthology.org/C18-1298} {Bridging resolution: Task definition, corpus resources and rule-based experiments}.
\newblock In \emph{Proceedings of the 27th International Conference on Computational Linguistics}, pages 3516--3528, Santa Fe, New Mexico, USA. Association for Computational Linguistics.

\bibitem[{Subramanian and Roth(2019)}]{subramanian-roth-2019-improving}
Sanjay Subramanian and Dan Roth. 2019.
\newblock \href {https://doi.org/10.18653/v1/S19-1021} {Improving generalization in coreference resolution via adversarial training}.
\newblock In \emph{Proceedings of the Eighth Joint Conference on Lexical and Computational Semantics (*{SEM} 2019)}, pages 192--197, Minneapolis, Minnesota. Association for Computational Linguistics.

\bibitem[{Toshniwal et~al.(2021)Toshniwal, Xia, Wiseman, Livescu, and Gimpel}]{toshniwal-etal-2021-generalization}
Shubham Toshniwal, Patrick Xia, Sam Wiseman, Karen Livescu, and Kevin Gimpel. 2021.
\newblock \href {https://doi.org/10.18653/v1/2021.crac-1.12} {On generalization in coreference resolution}.
\newblock In \emph{Proceedings of the Fourth Workshop on Computational Models of Reference, Anaphora and Coreference}, pages 111--120, Punta Cana, Dominican Republic. Association for Computational Linguistics.

\bibitem[{Weischedel et~al.(2011)Weischedel, Hovy, Marcus, Palmer, Belvin, Pradhan, Ramshaw, and Xue}]{weischedel-handbook-2011-notes}
Ralph Weischedel, Eduard Hovy, Mitchell Marcus, Martha Palmer, Robert Belvin, Sameer Pradhan, Lance Ramshaw, and Nianwen Xue. 2011.
\newblock {O}nto{N}otes: A large training corpus for enhanced processing.
\newblock In Joseph Olive, Caitlin Christianson, and John McCary, editors, \emph{Handbook of Natural Language Processing and Machine Translation: DARPA Global Autonomous Language Exploitation}. Springer.

\bibitem[{Wu et~al.(2020)Wu, Wang, Yuan, Wu, and Li}]{wu-etal-2020-corefqa}
Wei Wu, Fei Wang, Arianna Yuan, Fei Wu, and Jiwei Li. 2020.
\newblock \href {https://doi.org/10.18653/v1/2020.acl-main.622} {{C}oref{QA}: Coreference resolution as query-based span prediction}.
\newblock In \emph{Proceedings of the 58th Annual Meeting of the Association for Computational Linguistics}, pages 6953--6963, Online. Association for Computational Linguistics.

\bibitem[{Wu and Gardner(2021)}]{wu-gardner-2021-understanding}
Zhaofeng Wu and Matt Gardner. 2021.
\newblock \href {https://doi.org/10.18653/v1/2021.crac-1.16} {Understanding mention detector-linker interaction in neural coreference resolution}.
\newblock In \emph{Proceedings of the Fourth Workshop on Computational Models of Reference, Anaphora and Coreference}, pages 150--157, Punta Cana, Dominican Republic. Association for Computational Linguistics.

\bibitem[{Yu and Poesio(2020)}]{yu-poesio-2020-multitask}
Juntao Yu and Massimo Poesio. 2020.
\newblock \href {https://doi.org/10.18653/v1/2020.coling-main.315} {Multitask learning-based neural bridging reference resolution}.
\newblock In \emph{Proceedings of the 28th International Conference on Computational Linguistics}, pages 3534--3546, Barcelona, Spain (Online). International Committee on Computational Linguistics.

\bibitem[{Zeldes(2017)}]{Zeldes2017}
Amir Zeldes. 2017.
\newblock \href {https://doi.org/http://dx.doi.org/10.1007/s10579-016-9343-x} {The {GUM} corpus: Creating multilayer resources in the classroom}.
\newblock \emph{Language Resources and Evaluation}, 51(3):581--612.

\bibitem[{Zhu et~al.(2021)Zhu, Pradhan, and Zeldes}]{zhu-etal-2021-ontogum}
Yilun Zhu, Sameer Pradhan, and Amir Zeldes. 2021.
\newblock \href {https://doi.org/10.18653/v1/2021.acl-short.59} {{O}nto{GUM}: Evaluating contextualized {SOTA} coreference resolution on 12 more genres}.
\newblock In \emph{Proceedings of the 59th Annual Meeting of the Association for Computational Linguistics and the 11th International Joint Conference on Natural Language Processing (Volume 2: Short Papers)}, pages 461--467, Online. Association for Computational Linguistics.

\end{thebibliography}
\bibliographystyle{acl_natbib}

\appendix

\section{Model architecture oveview}
\label{sec:appendix_modelarc}
\begin{figure*}[t!hb]
    \centering
    \includegraphics[clip,width=1\textwidth,trim = 0cm -2cm 0cm 0cm]{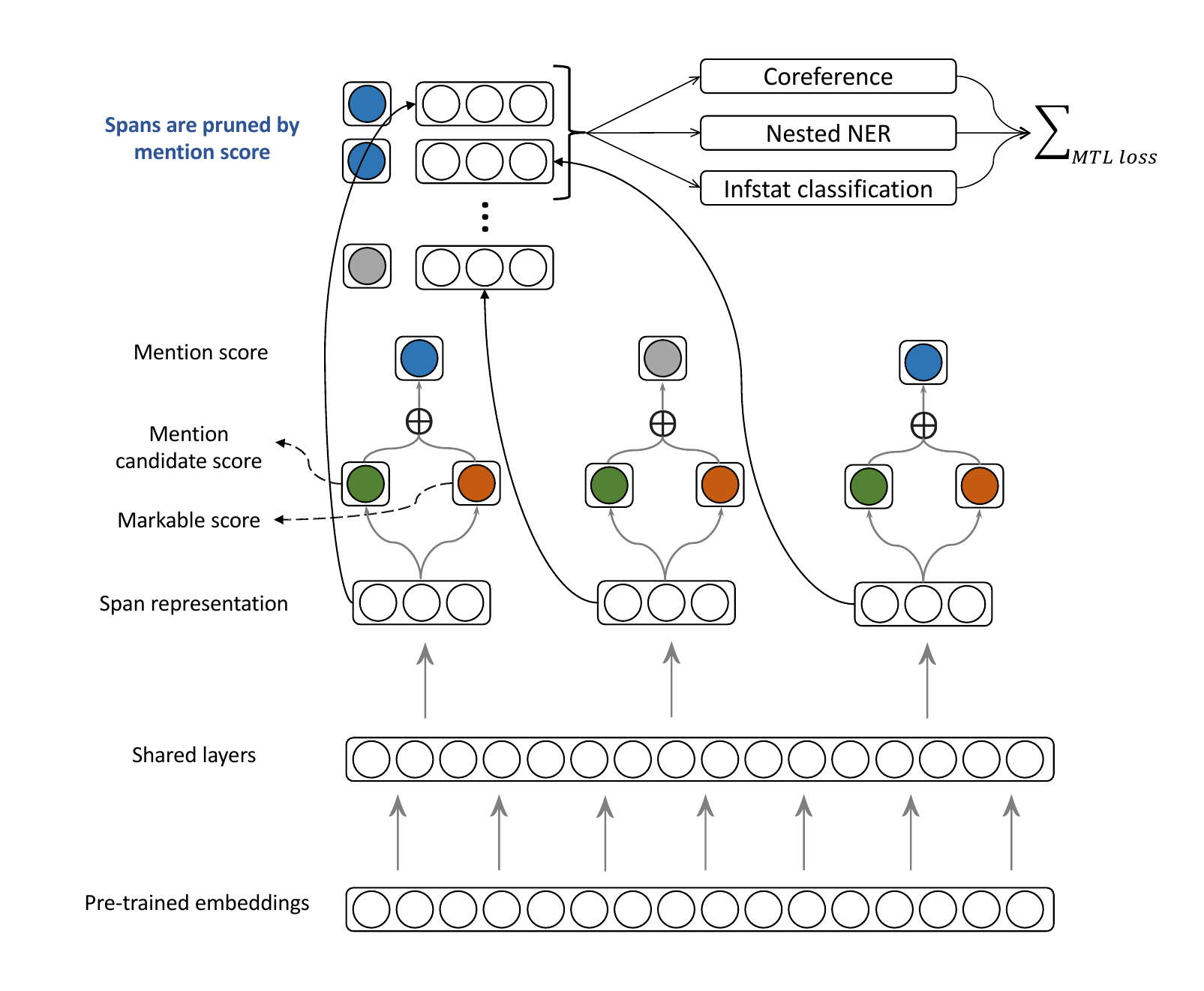}
    \caption{An overview of the proposed MTL model architecture. Only selected spans with high mention scores (in blue) are considered in the three auxiliary tasks.} \label{fig:model}
\end{figure*}

Figure \ref{fig:model} shows the architecture of the model proposed in this paper.

\section{Implementation details}
\label{sec:appendix_implementation}
We use Pytorch and the pre-trained SpanBERT-large \citep{joshi-etal-2020-spanbert} model from HuggingFace\footnote{\url{https://huggingface.co/}} for token representations. Experiments run on Nvidia RTX A6000 GPUs with 64GB RAM. Following previous work \citep{lee-etal-2018-higher, joshi-etal-2020-spanbert}, we use a batch size of 1 document for training and evaluation. The coreference task uses the same loss strategy as the baseline model \citep{joshi-etal-2020-spanbert} and each auxiliary task uses Cross Entropy loss. We use AdamW to optimize coreference loss and Adam to optimize auxiliary loss. We train 14,500 steps with \texttt{task\_learning\_rate} of 0.0003 for baselines and our models.

\section{Information Status}
\label{sec:appendix_infstat}
There are six types of information status in the data: 

\begin{itemize}[itemsep=0pt]
    \item \textit{new} (first, unmediated mention of an entity)
    \item \textit{given:active} (subsequent mention after a recent previous mention)
    \item \textit{given:inactive} (subsequent mention of a non-recently mentioned entity)
    \item \textit{accessible:inferrable} (new entity whose existence could be inferred from other mentions, e.g.~via bridging anaphora \cite{roesiger-etal-2018-bridging,hou-2020-bridging}, as in \textit{a house ... [the door]})
    \item \textit{accessible:commonground} (entities accessible to speakers in the situation, e.g.~\textit{pass [the salt]!})
    \item \textit{accessible:aggregate} (new entities referring back to multiple entities, i.e.~split antecedents as in \textit{Kim ... Yun ... [they]}). 

\end{itemize}

\noindent When information status is included in the auxiliary tasks, our model is trained to predict the label for each mention.

\section{Sample data}
\label{sec:appendix_sample}

\begin{figure*}[t!bh]
\centering
\begin{subfigure}[b]{0.80\textwidth}
\centering
\includegraphics[clip,width=1\textwidth,trim = 0cm 10cm 0cm 0cm]{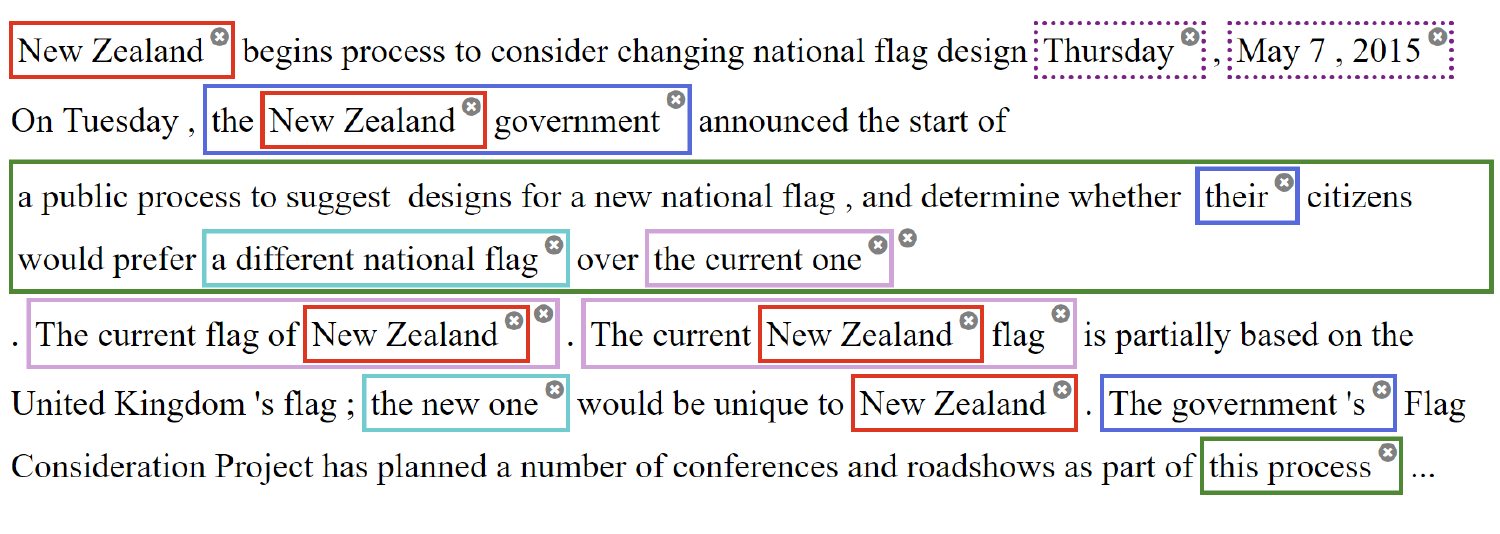}
\caption{Information available to the baseline e2e model.}
\label{fig:og-all-baseline}
\end{subfigure}
\hfill
\begin{subfigure}[b]{0.80\textwidth}
\centering
\includegraphics[clip,width=1\textwidth,trim = 0cm 2cm 0cm 0cm]{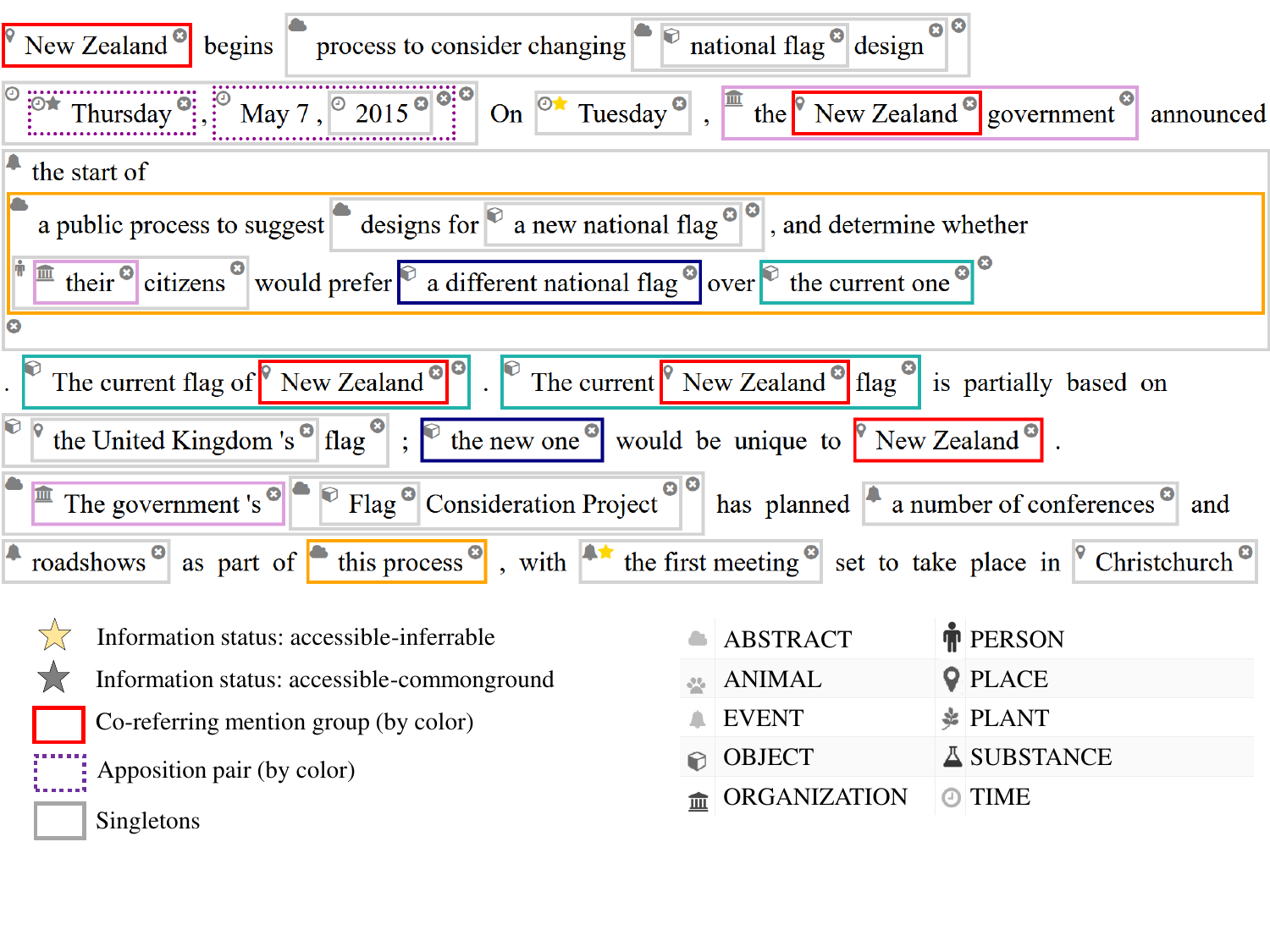}
\caption{Information available to the MTL model for the same document.}
\label{fig:og-all-mtl}
\end{subfigure}
\caption{Training data from an OntoGUM article in the news genre.}
\label{fig:og-all}
\end{figure*}

Figure \ref{fig:og-all-mtl} shows the extent of annotations available to the MTL model for training, compared to the data restricted to coreferring pairs in Figure \ref{fig:og-all-baseline}, as used by the baseline e2e approach. Since OntoNotes-style data, such as OntoGUM, does not contain singletons, mention types or information status, only coreferring mentions and their spans can be used for learning by the baseline model. The information in the bottom panel, by contrast, is much richer and covers all referring expressions with information status and one of ten entity types: \textsc{abstract, animal, event, object, organization, person, place, plant, substance} and \textsc{time}.

While each information type (coreference, mention boundaries, mention types, and information status) is not totally predictable from others, they overlap to some extent and exhibit different information densities: mention boundaries are available for many spans and are densely attested. Mention types are available for each mention, but some types are rare, e.g.~abstract mentions marked by \textcolor{gray}{\faCloud} in Figure \ref{fig:og-all} are the most common. Information status is mostly predictable from coreference, e.g.~singletons and chain-initial mentions are \textit{new}, and chain-medial or final mentions are given (\textit{given:active} if recently mentioned, otherwise \textit{given:inactive}). Accessible mentions are less trivial and comparatively rare (about 7.1\% of mentions in GUM), indicating whether they are accessible in the common ground, their identity is inferrable from some other mention (\textit{accessible:inferrable}), or by aggregating information from multiple mentions (\textit{accessible:aggregate}). This information could help systems to learn whether a span is likely to have an antecedent.

\end{document}